\documentclass[conference,10pt]{IEEEtran}

\usepackage[T1]{fontenc}
\usepackage{color}
\usepackage{psfrag}
\usepackage[dvips]{graphicx}
\usepackage{tikz}
\usepackage{pgfplots}
\pgfplotsset{compat=1.12}
\usepackage{enumerate}
\usepackage{rotating}
\usepackage{srcltx}
\usepackage{booktabs}
\usepackage{multicol}
\usepackage{enumitem}
\usepackage{textcomp}
\usetikzlibrary{shapes,arrows}
\usetikzlibrary{arrows,decorations.pathmorphing,backgrounds,positioning,fit,matrix}

\usepackage{csquotes}

\usepackage[noadjust]{cite}
\usepackage[cmex10]{amsmath}
\usepackage{amsfonts}
\usepackage{amssymb}
\usepackage{dsfont}    %mathds offers better set-symbols for N,Z,Q,R,C
\usepackage{mathtools} %\coloneqq for perfect :=
\usepackage{stackengine}

\definecolor{PU_orange}{HTML}{EE7F2D}
\definecolor{PU_darkorange}{HTML}{994400}
\definecolor{PU_lightorange}{HTML}{FFAA66}
\definecolor{PU_black}{HTML}{000000}
\definecolor{PU_darkgray}{HTML}{7F7F83}
\definecolor{PU_lightgray}{HTML}{BDBEC1}

\usepackage{algorithmic}
\usepackage{array}
\usepackage[caption=false,font=footnotesize]{subfig}

\newcommand{\E}{\mathbb{E}}

\newcommand{\bbE}{\mathbb{E}}\newcommand{\rme}{\mathrm{e}}

\newcommand{\bbR}{\mathbb{R}}

\newcommand{\bfS}{\mathbf{S}}

\newcommand{\cA}{\mathcal{A}}

\newcommand{\cI}{\mathcal{I}}

\newcommand{\cN}{\mathcal{N}}

\newcommand{\cS}{\mathcal{S}}

\newcommand{\cW}{\mathcal{W}}

\newcommand{\mfrI}{\mathfrak{I}}

\usepackage{comment}
\usepackage{soul}

\makeatletter
\g@addto@macro\normalsize{%
  \setlength\abovedisplayskip{2pt}
  \setlength\belowdisplayskip{2pt}
  \setlength\abovedisplayshortskip{1pt}
  \setlength\belowdisplayshortskip{1pt}
}
\makeatother

\usepackage{amsthm}
\newtheoremstyle{mystyle}% name
{}% space above
{}% space below
{\itshape}% body font
{}% indent amount
{\bfseries}% theorem head font
{}% punctuation after theorem head
{.5em}% space after theorem head
{}% theorem head spec

\newtheoremstyle{remark}% name
{}% space above
{}% space below
{}% body font
{}% indent amount
{\itshape}% theorem head font
{}% punctuation after theorem head
{.5em}% space after theorem head
{}% theorem head spec

\makeatletter
\def\thmhead@plain#1#2#3{%
  \thmname{#1}\thmnumber{\@ifnotempty{#1}{ }\@upn{#2.}}%
  \thmnote{ \small\textsf{\the\thm@notefont\textit{#3}.}}}
\let\thmhead\thmhead@plain
\makeatother

\theoremstyle{mystyle}
\newtheorem{theorem}{Theorem}%[section]
\theoremstyle{mystyle}
\newtheorem{lemma}{Lemma}%[section]
\theoremstyle{mystyle}
\theoremstyle{mystyle}
\theoremstyle{mystyle}
\newtheorem{definition}{Definition}%[section]
\theoremstyle{remark}
\theoremstyle{mystyle}
\theoremstyle{mystyle}
\newtheorem{exa}{Example}%[section]
\theoremstyle{mystyle}
\theoremstyle{discussion}
\theoremstyle{mystyle}
\theoremstyle{mystyle}
\usepackage{pgf,tikz,pgfplots}
\pgfplotsset{every tick label/.append style={font=\fontsize{1}{2}\selectfont}}
\usepackage{mathrsfs}
\usetikzlibrary{arrows}
\allowdisplaybreaks

\long\def\symbolfootnote[#1]#2{\begingroup%
	\def\thefootnote{\fnsymbol{footnote}}\footnote[#1]{#2}\endgroup} %unnumbered footnote for thanks
\IEEEoverridecommandlockouts    
\title{Information-Theoretic Bounds on the Generalization Error and Privacy Leakage in Federated Learning} 
\author{\IEEEauthorblockN{Semih Yagli, Alex Dytso,  H. Vincent Poor,}
Princeton University, Princeton, NJ 08544, USA, Email: \{syagli, adytso, poor\}@princeton.edu
\thanks{Accepted for publication in Proceedings of 21st IEEE International Workshop on Signal Processing Advances in Wireless Communications (SPAWC). This work was supported in part by the U. S. National Science Foundation under Grant CCF-1908308, and in part by the Princeton Center for Statistics and Machine Learning under a DataX grant.}}

\begin{document}%
\maketitle

\begin{abstract}
Machine learning algorithms operating on mobile networks can be characterized into three different categories. First is the classical situation in which the end-user devices send their data to a central server where this data is used to train a model. Second is the distributed setting in which each device trains its own model and send its model parameters to a central server where these model parameters are aggregated to create one final model. Third is the federated learning setting in which, at any given time $t$, a certain number of active end users train with their own local data along with feedback provided by the central server and then send their newly estimated model parameters to the central server. The server, then, aggregates these new parameters, updates its own model, and feeds the updated parameters back to all the end users, continuing this process until it converges.

The main objective of this work is to provide an information-theoretic framework for all of the aforementioned learning paradigms. Moreover, using the provided framework, we develop upper and lower bounds on the generalization error together with bounds on the privacy leakage in the classical, distributed and federated learning settings. 

\end{abstract}

\section{Introduction}

 Introduced in \cite{mcmahan2016communication} as a model for studying  decentralized learning over mobile networks, the \emph{federated learning paradigm}  takes a decentralized approach to machine learning problems. The objective of federated learning is to collaboratively learn a shared global model from the participating end-user devices, which retain control of their own data. While it is especially useful in situations where there is limited communication bandwidth in user-server interaction, distributing the learning task also helps retain the privacy of data for each user.
  
In federated learning, each user maintains its local dataset, and instead of  forwarding the entire data to a central node, each user computes  and sends a small update to the global model  maintained by the server.  Although the similarities may be more than pronounced, the classical distributed learning, e.g., \cite{shamir2014distributed}, differs from the federated learning in several ways in that the latter provides more flexibility. For example, unlike the former, the datasets in the latter are not necessarily balanced. Moreover, unlike in distributed learning, the data in federated learning are not assumed to be generated by independent and identically distributed (i.i.d.) random variables.

The objectives of this work are two-fold. The first  is to develop bounds on the \emph{generalization error} of a general federated learning algorithm. A non-computable quantity on its own, the generalization error measures  the accuracy of an algorithm in its prediction of the outcomes for previously unobserved data. This, in turn, quantifies how much  the learning algorithm overfits its training data.  The second objective is to quantitatively measure the inherent \emph{privacy} offered by the distributed nature of federated learning. We  meet both of these objectives using an information-theoretic approach.
 
In regard to the recent literature,  for the classical setting where the server has  access to all of the data, there have been several results on bounding the generalization error using information-theoretic measures. The  initial   work in this direction was due to Russo and Zuo  in \cite{russo2015much} where they used mutual information  to establish bounds on the generalization error of a learning algorithm.   The work of \cite{russo2015much} was  later generalized in \cite{jiao2017dependence} and \cite{xu2017information} to include  infinitely many hypotheses. Some of the bounds proposed in \cite{jiao2017dependence} and \cite{xu2017information} were further tightened in \cite{asadi2018chaining} and \cite{bu2019tightening}. The idea of using mutual information to bound generalization error has also been extended to noisy iterative algorithms, cf. \cite{bu2019tightening, pensia2018generalization, negrea2019information}.  While  bounds on the generalization error in terms of the Wasserstein distance can be found in \cite{raginsky2016information} and \cite{wang2019information}, bounds on this sought-after in terms of other  information-theoretic metrics such as $f$-divergences and $\alpha$-mutual information has been respectively considered in \cite{jiao2017dependence} and \cite{esposito2019generalization}.

A standard approach to measuring the privacy of the user data is through the notion of differential privacy, which was introduced in \cite{dwork2006calibrating}.  In this work, in place of the pure differential privacy, we  prefer a  different yet strongly related notion, namely the \emph{Bayesian mutual information privacy}, cf. \cite{cuff2016differential} and \cite{yang2015bayesian}. The main reason behind the choice of this metric over the $\epsilon$-differential privacy is because the former is more amenable to computations.

Federated learning has recently received considerable attention in the literature. Due to space limitation we do not seek to survey the work in this area, and instead,  for recent surveys the interested reader is referred to \cite{li2019federated} and \cite{kairouz2019advances}. 

As for the organization, Section~\ref{sec:DefinitionOfLearninModels} defines and differentiates the three learning paradigms: classical (or centralized) learning, distributed learning, and, last but not least, federated learning.   Section~\ref{sec:InforTheoreAndPrivacy} overviews  information-theoretic notions related to generalization bounds and differential privacy and Section~\ref{sec:MainResults} presents our main results and demonstrates several examples.

\section{On Various Learning Models}
\label{sec:DefinitionOfLearninModels}
%In this section, we review learning models.  

\subsection{Centralized Learning} 
A classical learning setup, referred here as the centralized learning problem, can be defined as follows. 

\begin{definition}[Centralized Learning]\label{def:ClassicalLearningAlgorithm} A centralized learning problem consists of the following: 
\begin{itemize}[leftmargin=*]
\item   an instance space $\mathcal{S}$, and a hypothesis space $\mathcal{W}$;
\item   a nonegative loss function $\ell:\mathcal{S} \times \mathcal{W} \to \mathbb{R}^{+}$;
\item   an input dataset  of size $n$, that is:   
\begin{align*}
\mathbf{S}=(S_{1}, \ldots, S_{n}),   
\end{align*}
 where  elements of $\mathbf{S}$ are generated i.i.d. according to some  prior distribution $\pi$.   The distribution of  the dataset $\mathbf{S}$ is denoted by $P_{\mathbf{S}}= \pi^{ \otimes n } $; and 
\item  a learning algorithm characterized by a conditional  probability distribution $P_{W|\mathbf{S}} $.  A learning algorithm  takes a dataset $\mathbf{S}$ and outputs  $W \in \mathcal{W}$ according to  $ P_{W|\mathbf{S}}$. We denote this learning algorithm by $\mathcal{A}$. 
\end{itemize} 
\end{definition} 

The objective of the learning problem is to make sure that for a chosen $W$ the \emph{expected risk} (or \emph{true error}) is small.  The risk of a hypothesis $w \in \mathcal{W}$ with respect to the prior distribution  $P_{S} = \pi$ is denoted by 
\begin{align*}
L_{\pi}(w)= \E\left[ \ell(S,w) \right]. 
\end{align*} 
As the prior $\pi$ is typically unknown, it is impossible to estimate the true error. For this reason, the \emph{empirical risk} over the observed dataset $\mathbf{S}$ is often used as a substitute: % which is defined as
\begin{align*}
L_{\mathbf{S}}(w)= \frac{1}{n} \sum_{i=1}^{n} \ell( S_i,w).  
\end{align*}

% Some examples of classical learning tasks can be listed as follows.
%
%\begin{exa}[Binary Classification]  The instance space is given by   $\mathcal{S}= \mathcal{X} \times \mathcal{Y}$ where $\mathcal{X}$ is the domain set  and  $\mathcal{Y}=\{0,1 \}$ is the set of labels.    The hypothesis space  $\mathcal{W}$ is given by  the set of binary classifiers $w: \mathcal{X} \to \mathcal{Y}$.  Finally, the loss function is  given  by zero-one loss as  $\ell(s,w)=\mathsf{1}_{ \{w(x) \neq y \} }$. 
%\end{exa} 
%
%\begin{exa}[Linear Regression]    The instance space is given by   $\mathcal{S}= \mathcal{X} \times \mathcal{Y}$ where  the domain set $\mathcal{X} \subset \mathbb{R}^n$ for some $n$,  and  the set of labels $\mathcal{Y}=\mathbb{R}$. The hypothesis space  $\mathcal{W}$ is given by the set of all linear functions $ \mathcal{W}=\{ w= \langle a, x \rangle +b:  a \in \mathbb{R}^n, b \in \mathbb{R}   \} $. 
%\end{exa} 

For an algorithm $P_{W|\mathbf{S}}$, the \emph{generalization error} is defined as the difference  between the true error and the empirical risk, namely $L_{\pi}(W) - L_{\mathbf{S}}(W)$.  Useful in many contexts, the  \emph{average}  (with respect to $W \sim P_{W} $) of this quantity is %defined as
\begin{align*}
{\rm gen}(\pi, \mathcal{A})=  \E \left[     L_{\pi}(W)- L_{\mathbf{S}}(W)   \right].  
\end{align*} 

% Due to the multi-sample nature of our problem setting, to asses the sensitivity of the generalization error to a single sample $j$,  we define the following notion,  which we call the \emph{sensitivity}  of the algorithm $\cA $ on the sample $j$: 
%\begin{align*} 
%{\rm Sen} ( \pi, \mathcal{A}; j ) =    \E \left [ \left | \E \left[   L_{\pi}(W)- L_{\mathbf{S}}(W)  | \mathbf{S}^{-j}  \right]  \right|^2   \right]  ,
%\end{align*} 
%where the expectation is taken with respect to $P_{W|\mathbf{S}} P_{\mathbf{S}}$.  The sensitivity, ${\rm Sen}(\pi, P_{W|\mathbf{S}}; j)$, measures the degree at which the output of an  algorithm $P_{W|\mathbf{S}}$ depends on a particular example, say $j$, in the data set.  As an immediate property,  by Jensen's inequality, we have  for  any arbitrary  $j$
%\begin{align}
%|{\rm gen}(\pi,\mathcal{A})| \le \sqrt{  {\rm Sen} ( \pi, \mathcal{A}; j ) }.  \label{eqn:gen<sen}
%\end{align} 
%In other words, as expected,  the more the average generalization error,  the more the  sensitivity  for each data entry. 
%

\subsection{Distributed Learning Without Interaction}
In many modern applications  especially the ones involving mobile phones, tablets, and laptops, the datasets are no longer centralized but rather are distributed amongst multiple users.  

 Describing a better model for such applications,  this section presents a general mode for distributed learning where the datasets are stored locally on user devices. In addition to capturing modern features of a distributed algorithm such as the possibility that each user  can have different number of data points \cite{mcmahan2016communication}, the proposed definition down below  also incorporates  an increasingly  important feature of data being  generated according to  distinct  probability distribution for each user \cite{mcmahan2016communication}. Besides its flexibility, the  below definition allows the possibility of randomized algorithms  as well. 

\begin{definition}[Distributed Learning]\label{def:distributedLearningNoInteraction}
A distributed learning problem  with $K$ users and a single central server consists of the following:
\begin{itemize}[leftmargin=*]
\item   an instance space $\mathcal{S}$, and a hypothesis space $\mathcal{W}$;
\item   a nonegative loss functions $\ell:\mathcal{S} \times \mathcal{W} \to \mathbb{R}^{+}$;
\item   $K$ input datasets each of size $n_k, k \in [1:K]$, that is  
\begin{align*}
\mathbf{S}_k=(S_{k,1}, \ldots, S_{k,n_k}),   
\end{align*}
where  $S_{k,j} \in  \mathcal{S}, \forall k,j$  and    each  are generated i.i.d. according to some distribution $\pi_k$.  Moreover, we assume that $\{\mathbf{S}_k \}_{k=1}^K$ are independent.  From here on, the set $ \mathbf{S}_K= \bigotimes_{k=1}^K   \mathbf{S}_k$  shall be referred to as  the \emph{global training set};

\item  a set of  $K$ local learning algorithms each characterized by the conditional probabilities  $ \{   P_{W_k| \mathbf{S}_k} \}_{k=1}^K $.  Here, the learning algorithm $k$ takes  the dataset $ \mathbf{S}_k$ and  outputs $W_k \in \mathcal{W}$ randomly according to  $ P_{W_k| \mathbf{S}_k}$;  and
\item   a fusion algorithm characterized by the conditional probability distribution $P_{\widehat{W}|W_1,\ldots, W_K}$ that randomly assigns an estimate $\widehat{W} \in \mathcal{W}$  based on the observed  $K$-tuple $(W_1, \ldots, W_K)$. 
\end{itemize}
We refer to the collection of local learning procedures, $ \{P_{W_i| \mathbf{S}_i} \}_{i=1}^K $, and the fusion procedure, $P_{\widehat{W}|W_1,\ldots, W_K}$, as the \emph{distributed learning algorithm} and denote the whole collection by $\mathcal{A}_K = (\{P_{W_i| \mathbf{S}_i} \}_{i=1}^K , P_{\widehat{W}|W_1,\ldots, W_K}) $. 
\end{definition}

The expected risk for the distributed learning algorithm for a hypothesis $\widehat{w} \in \mathcal{W}$ is denoted by 
\begin{align*}
L_{ \mu } (\widehat{w})=   \sum_{k=1}^K \frac{n_k}{n} L_{\pi_k} (\widehat{w}) = \E \left[  \sum_{k=1}^K \frac{n_k}{n}  \ell(S_k,\widehat{w})   \right],
\end{align*}  
where  $n= \sum_{k=1}^K n_k$ is the total number of training examples,  $S_k \sim \pi_k$  and the expectation is taken with respect to the product of priors $\mu = \prod_{k=1}^K \pi_k $.  Observe that, to account for the  unequal  amount of data that each user has,  we weight the population risk accordingly.

 Similarly, the empirical risk for a hypothesis $\widehat{w} \in \mathcal{W}$ is %defined as 
\begin{align*}
L_{\mathbf{S}_K }  (\widehat{w})= \sum_{k=1}^K  \frac{n_k}{n}  L_{\mathbf{S}_k}(\widehat{w}).
\end{align*}

For a distributed learning algorithm, the  expected generalization error  is defined by 
\begin{align*}
{\rm gen}(\mu;    \mathcal{A}_K)  =  \E \left[     L_{\mu}( \widehat{W})- L_{\mathbf{S}_K}(\widehat{W})   \right], %\label{eq:defGeneralizationError}
\end{align*} 
%and the sensitivity is defined as
%\begin{align*}
% {\rm Sen} ( \pi, \mathcal{A}_K; j_m ) =\E \left[  \left |   \E \left[  L_{\mu}( \widehat{W})- L_{\mathbf{S}_K}(\widehat{W})  \middle | \mathbf{S}_{K }^{-j_m} \right ]  \right |^2   \right]. 
%\end{align*} 
%Similar to its classical counterpart in \eqref{eqn:gen<sen},  the following inequality between generalization error and sensitivity  holds:
%\begin{align*}
% \left| {\rm gen}(\mu;     \mathcal{A}_K) \right| 
% \le  \sqrt{  {\rm Sen} ( \pi, P_{W|\mathbf{S}}; j_m ) }. 
%\end{align*} 

 Below we list some examples of  distributed learning tasks.
\begin{exa}[Classical Learning]\label{example:Distribute_centr}
Setting $W_i= \mathbf{S}_i$  reduces to the case when the central unit has access to all of the datasets. Moreover, the Definition~\ref{def:distributedLearningNoInteraction} reduced to the classical setting by further assuming that the datasets have the same prior distribution. 
\end{exa}  

Sharing  each and all of the datasets, as in Example~\ref{example:Distribute_centr},  with the central unit  incurs a very high communication cost. Therefore, instead of uploading their respective datasets, it would be preferable if users  could perform local training.  This way, they would learn their own  models  from their local training  and upload those learnt models to the central user.  Then, upon receiving the models, the central user would  apply a fusion rule to aggregate the models and  create one global model. A simple example of this procedure is illustrated next.  
\begin{exa}[Model Aggregation]\label{Example:ModelAggreg}  Suppose that the loss function is given by $\ell(S,w)= \|S-w \|^2$ where $\| \cdot\|$ is the Euclidian norm.   Instead of sending the entire dataset $\mathbf{S}_k$, user $k$ performs local empirical risk minimization, i.e.,  minimizes $L_{\mathbf{S}_k}(w)=\frac{1}{n_k} \sum_{i=1}^{n_k} \|S_{k,i}-w \|^2$.   Then, this  minimum, which  is given by a sample mean  $W_i= \frac{1}{n_k} \sum_{i=1}^{n_k} S_{k,i}$, is sent to the central user.     The central user performs model aggregation by, for example, performing the weighted average 
\begin{align*}
\widehat{W}= \sum_{k=1}^K \frac{n_k}{n} W_k. 
\end{align*} 
The generalization error and privacy leakage for this example are computed in Lemma~\ref{lem:MeanEstiamtionDistributeSetting}. 
\end{exa}  

 Apart from the communication efficiency, an additional benefit of  sending the model instead of the whole dataset is in terms of the privacy of user's data. Thanks to its definition in Definition~\ref{def:distributedLearningNoInteraction}, there already is an inherent privacy offered by the distribution of the learning task. Presented in Section~\ref{sec:BOundsDecentralizedCase}, Lemma~\ref{lem:MeanEstiamtionDistributeSetting} gives an illustration of this.

\subsection{Federated Learning} 

An idea much similar to distributed learning, federated learning is born to meet the needs of our ultra-connected modern world. Similar to the distributed learning, in the federated learning setup, users, i.e., phones, tablets, laptops and other IoT devices, train their own models by using their own data, and send the resulting parameters of their trainings to a central server. However, unlike in distributed learning, the central server does not expect this training to be done in just a single round. The replacing assumption in federated learning is that after a certain number\footnote{Not all users are expected to be active at all times.} of users train and send their model parameters to the central server, the central server aggregates and informs the global model parameters back to every user. Upon receipt of global model parameters and more data to work on, active users now re-estimate their model parameters and send their new updates to the central server, continuing this process until convergence.  

Note the key differences between the federated and distributed learning schemes. First and foremost, in federated learning, there are several communication rounds where the central server interacts with a certain subset of users. Moreover, in federated learning, the central server does not assume that every user is active at all times. Not only does this provide a better model for our daily use of IoT devices, but it also helps federated learning be more robust to single node failures.

%
%\st{While  Definition}~\ref{def:distributedLearningNoInteraction}\st{ captures many of the distribute learning scenarios, it is still  missing two main components. The first component is a possibility of several rounds of communications where the central user is allowed to interact with all or a subset of users.  The second component is a possibility of random sampling of users and subset of user data. Incorporation of these two components into Definition}~\ref{def:distributedLearningNoInteraction}\st{ leads to the paradigm that is known as federated learning. }
%

\begin{definition}[Federated Learning]\label{def:FederatedLearningAlgorithm}
A  federated learning problem  with $K$ users and one central  server consists of the following:
\begin{itemize}[leftmargin=*]
\item   an instance space $\mathcal{S}$, and a hypothesis space $\mathcal{W}$;
\item   a nonnegative loss functions $\ell:\mathcal{S} \times \mathcal{W} \to \mathbb{R}^{+}$;
\item  a timing set $\{1,\ldots, T \}$ where $T$ is the total number of communication rounds; 
\item   $K$ input datasets  $\left\{\mathbf{S}_k\right\}_{k=1}^K$.  The dataset  $\mathbf{S}_k$ is divided into $T$ subsets (batches) each of size $n_k, k \in [1:K]$, that is  
\begin{align*}
\mathbf{S}_k=(\mathbf{S}_{k}^{(1)},  \ldots, \mathbf{S}_{k}^{(T)}),   
\end{align*}
where for  $ t\in \{1,\ldots, T \}$ 
\begin{align*}
\mathbf{S}_{k}^{(t)}=( S_{k, 1}^{(t)}, \ldots, S_{k, n_k}^{(t)} ), 
\end{align*} 
with  $S_{k,j}^{(t)} \in  \mathcal{S}, \forall k, j, t$  and each  are generated i.i.d.  according to some  distribution $\pi_k$.  Moreover, we assume that $\{\mathbf{S}_k \}_{k=1}^K$ are independent.  From here on, the set $ \mathbf{S}_K= \bigotimes_{k=1}^K   \mathbf{S}_k$  shall be referred to as  the \emph{global training set;}
\item   a random sample      $\mathcal{I}^{(t)} \subseteq \{1, ..., K\}$   of  $K_a^{(t)}$  active  users chosen at  time $t$;  
\item  a set of  $K$ iterative learning algorithms each characterized by the conditional probabilities $ \left\{P_{W_k^{(t)}| \mathbf{S}_{k}^{(t)}, \widehat{W}^{(t-1)} }\right\}_{k=1}^K. $ Here, the learning algorithm  at time $t$  takes  the dataset (batch) $ \mathbf{S}_{k}^{(t)}$ and  the previous output of the fusion center  $\widehat{W}^{(t-1)}$ (defined below) and  outputs $W_k^{(t)} \in \mathcal{W}$ at time $t$   according to   $P_{W_k^{(t)}| \mathbf{S}_{k}^{(t)}, \widehat{W}^{(t-1)} }$; and 
\item  a (possibly) randomized output $\widehat{W}^{(t)}$  of a fusion algorithm that depends on the outputs of active users  at time $t$, namely $ \{ W_{i}^{ (t)} \}_{i \in \mathcal{I}^{(t)} } $.  
\end{itemize}
We refer to the collection of local learning and fusion procedures as the \emph{federated learning algorithm} and denote it by  $\mathcal{A}_K^{(T)}$.
\end{definition}

Akin to its distributed learning counterpart, the expected risk at time $t$ of the federated learning algorithm for a hypothesis $\widehat w \in \cW $ is denoted by
\begin{align*}
L^{(t)}_\mu(\widehat w; \cI^{(t)}) &= \sum_{k \in \cI^{(t)}} \frac{n_k}{n(t)} L_{\pi_k}(\widehat w) , 
% = \bbE\left[ \sum_{k \in \cI^{(t)}} \frac{n_k}{n(t)} \ell (S_k, \widehat w)\right]
\end{align*}
where $ n(t) = \sum_{i\in \cI^{(t)}} n_i  $ is the total number of available training examples at time $t$, $S_k \sim \pi_k $ and the expectation is taken with respect to the product of priors $\mu =  \prod_{k\in \cI^{(t)}} \pi_k $.

The empirical risk at time $t$ for a hypothesis $\widehat w \in \cW$ is defined as
\begin{align*}
 L^{(t)}_{\bfS_K} (\widehat w; \cI^{(t)}) = \sum_{k\in \cI^{(t)}} \frac{n_k}{n(t)} L_{\bfS_k^{(t)}}(\widehat w).
\end{align*}

For a federated learning algorithm, the expected generalization error at time $t$ is defined by 
\begin{align*}
 & {\rm gen}^{(t)} (\mu; \cA_K^{(T)}) \notag 
\\ &\ \ 
= \bbE \left[ \bbE\left[ L^{(t)}_\mu(\widehat W^{(t)}; \cI^{(t)}) -   L^{(t)}_{\bfS_K} (\widehat W^{(t)}; \cI^{(t)})  | \cI^{(t)}\right]\right] .   
\end{align*}

Note that the calculation of the expected generalization error is based on the difference between the expected risk and the empirical risk at the most current time $t$. The justification for this comes from the intuition behind the generalization error: as new samples arrive at time $t$, we would like to measure how our most up-to-date model does in fitting these new samples.

An example of the federated learning procedure can be demonstrated by adapting Example~\ref{Example:ModelAggreg} to this setting as shown next. 
\begin{exa}[Iterative Model Aggregation]\label{Example:ModelAggregFL} 
Let datasets and the loss function be as in Example~\ref{Example:ModelAggreg}. Consider the following evolution of the output of  the fusion center: 
\begin{align*}
\widehat{W}^{(t)}= \sum_{ k \in \mathcal{I}^{(t)}} \frac{n_k}{n^{(t)}} W_k^{(t)}. 
\end{align*}
That is,  $\widehat{W}^{(t)}$ is computed by aggregating outputs of randomly sampled users according to $\mathcal{I}^{(t)}$.  Moreover, the local decision  $\widehat{W}_K^{(t)}$ is computed by performing the following update:
\begin{align*}
	W_k^{(t)}=  \frac{1}{t n_k}  \sum_{j = 1 }^{n_k} S_{k,j}^{(t)}   + \frac{t-1}{t} \widehat{W}_K^{(t-1)},
	\end{align*}
	where  $W_k^{(0)} =\frac{1}{n_k} \sum_{i=1} S_{k,i}$. 
\end{exa}

%{\color{red} In the extended version give the Gradient Example} 

\section{Information Theoretic Preliminaries } 
\label{sec:InforTheoreAndPrivacy}

\subsection{Mutual Information} 

Given a correlated pair of random variables $(X, Y) \sim P_{XY}$, a fundamental information theoretical quantity called the \emph{mutual information} provides a symmetric measure of dependence between $X$ and $Y$ which is denoted by 
\begin{align*}
  I(X;Y) & =  D(P_{XY} \| P_XP_Y ) .
\end{align*}

The key to proving generalization bounds via mutual information, using Donsker-Varadhan variational representation of relative entropy and tools from duality theory, \cite[Theorem 1]{bu2019tightening} shows that it is possible to provide an upper bound, in terms of the mutual information, on the distance between the mean of a function under dependent random variables and the mean of the same function under independent random variables.

\begin{lemma}\label{lem:AuxiliaryConditionalLemma}

Let $(X,Y)\sim P_{XY} $, $(\bar X, \bar Y) \sim P_X P_Y $, and
\begin{align*}
	\Lambda_{f(\bar X, \bar Y)}(\lambda) = \log\bbE\left[\rme^{\lambda(f(\bar X, \bar Y) - \bbE[f(\bar X, \bar Y)])}\right] 
\end{align*}
denote the cumulant generating function of $f(\bar X, \bar Y)$. For $b_+ \in (0, \infty] $, if we can find a convex function $\psi_+ \colon [0, b_+) \to \bbR $ with $\psi_+(0) =  \psi_+'(0) = 0 $ satisfying  
\begin{align*}
  \Lambda_{f(\bar  X, \bar Y)}(\lambda) \le \psi_+(\lambda)  \text{ for } \lambda \in [0, b_+),
\end{align*}
then
\begin{align*}
 - \psi_+^{\ast-1}(I(X;Y)) \le    \bbE[f(\bar X, \bar Y)]  - \bbE[f(X,Y)]  ,
\end{align*}
 where $\psi_+^{\ast-1}$ denotes the inverse of the Legendre dual of $\psi_+$.

Similarly, for $b_- \in (0, \infty]$,  if we can find a convex function $\psi_- \colon [0, b_-)\to \bbR $ with $\psi_-(0) =  \psi_-'(0) = 0 $ satisfying
\begin{align*}
  \Lambda_{f(\bar  X, \bar Y)}(\lambda) \le \psi_-(-\lambda) \text{ for } \lambda \in (-b_-, 0],
\end{align*}
then
\begin{align*}
\bbE[f(\bar X, \bar Y)] - \bbE[f(X,Y)] \le  \psi_-^{\ast-1}(I(X;Y)). 
\end{align*}

\end{lemma}

%% OLD VERSION - OLD VERSION - OLD VERSION - OLD VERSION - OLD VERSION - OLD VERSION - OLD VERSION %%
%Recall the following definition.
%\begin{definition}
%A random variable $X$ is $\lambda$-sub-Gaussian if the following inequality holds:
%\begin{align*}
%\E \left[ \eu^{\lambda (X-\E[X])} \right ] \le \eu^{ \frac{\lambda^2 R^2}{2}},  \, \forall \lambda  \in \mathbb{R}
%\end{align*} 
%\end{definition} 
%
%We use the following lemma which is a conditional generalization of the bounds in   \cite[Lemma~1]{xu2017information}. 
%\begin{lemma}\label{lem:AuxiliaryConditionalLemma} Let $f(\bar{X}, \bar{Y})$ be $\lambda$-sub-Gaussian according to $P_{\bar{X},\bar{Y}  }=  P_{X| U=u   } P_{Y  |U=u }$ for all $u$. Then, 
%\begin{align*}
%\E \left [  \left | \E[f(X,Y)|U]  -  \E[f(\bar{X},\bar{Y})|U]  \right |^2   \right] \le \frac{\lambda^2}{2} I(X;Y|U). 
%\end{align*} 
%\end{lemma} 

\subsection{Privacy Leakage Measures} 

A standard in data science literature, differential privacy was introduced by Dwork \emph{et al.} in \cite{dwork2006calibrating}. 
%Although we use a information-theoretic variation of it, for the sake of completeness, the definition of $\epsilon$-differential privacy is given below.
%
%\begin{definition}[$\epsilon$-Differential Privacy] A randomized mechanism $P_{Y|\bfS} \colon \cS^n \to \cY$ satisfies $\epsilon$-Differential Privacy, $\epsilon$-DP, if for any adjacent $u^n$, $v^n \in \cS^n $ and  measurable set $\cD \subseteq \cY $
%\begin{align*}
%  \frac{P_{Y|\bfS= u^n}(\cD)}{P_{Y|\bfS = v^n}(\cD)} \in [\rme^{-\epsilon}, \rme^\epsilon] .
%\end{align*} 
%\end{definition}
In an effort to bring information-theoretic methods to study privacy, Cuff and Yu \cite{cuff2016differential} later introduced the so-called mutual information differential privacy ($\epsilon$-MIDP) which  can be defined as follows.

\begin{definition}[$\epsilon$-MIDP] 
  A randomized mechanism $P_{W|\bfS} \colon \cS^n \to \cW$ is said to be  $\epsilon$-mutual information differentially private, if 
  \begin{align*}
 \sup_\pi  \max_{i} I(S_i;W|S^{-i})  \le \epsilon,
\end{align*} 
where $S^{-i} = (S_1, \ldots, S_{i-1}, S_{i+1}, \ldots, S_n) $ and $(\bfS, W) \sim \pi^{\otimes n} P_{W|\bfS} $. 
\end{definition}

Inspired by the work of Cuff and Yu \cite{cuff2016differential}, throughout this paper we use the \emph{Bayesian mutual information differential privacy}, cf. \cite{yang2015bayesian}, where the \emph{privacy leakage} is defined via
\begin{align*}
   {\rm priv} (\pi, P_{Y|\bfS}) = \sup_{i} I(S_i;Y|S^{-i}). %\label{eqn:def:PrivacyLeakage}
\end{align*}

%\begin{definition}[$\epsilon$-MIDP]  Let  $\bfS=(S_1,\ldots, S_n)$ be a dataset  i.i.d. according to  $\pi$.  A randomized mechanism $P_{Y|\bfS} \colon \cS^n \to \cY$ is said to be  $\epsilon$-Mutual Information Private under the generation model $\pi$, $\epsilon$-MI-DP-$\pi$, if 
%\begin{align}
%{\rm priv}( P_{Y|\bfS})= \sup_{i} I(S_i;Y|S^{-i}) \le \epsilon \text{.}   \label{eq:DefPrivacy}
%\end{align}
%Further, we say that the mechanism is  $\epsilon$-Mutual Information Private, $\epsilon$-MI-DP, if 
%\begin{align}
%\sup_\pi {\rm priv}( P_{Y|\bfS})  \le \epsilon .
%\end{align} 
%\end{definition} 
%
%As our privacy leakage measure, we use
%\begin{align}
%  \sup_{i} I(S_i;Y|S^{-i}) 
%\end{align}
%\begin{rem}
%  As its name suggests, the lesser the bound on the privacy leakage ${\rm priv}( P_{Y|\bfS})$ the more the randomized algorithm provides privacy to its user. As a convention, if one can show that 
% \begin{align} 
%  {\rm priv}( P_{Y|\bfS}) = \infty,
% \end{align}
% this means that the randomized algorithm $P_{Y|\bfS}$ does NOT provide any privacy to its user.
%\end{rem}

%It can be shown that  \cite[Theorem~1]{cuff2016differential}
%\begin{align}
%\text{  $\epsilon$-MI-DP-$\pi$} \preceq  \text{  $\epsilon$-MI-DP}  \preceq \text{  $\epsilon$-DP} . 
%\end{align} 
%In this work we will mainly use $\epsilon$-MI-DP-$\pi$  which is weakest of the two definitions.

Note that the definition of privacy leakage %in \eqref{eqn:def:PrivacyLeakage} 
is well-tailored for the centralized case. However, the multi-user setup in distributed and federated learning problems calls for individual-based privacy leakage definitions. For this reason, for each user $k \in \{1, \ldots, K\}$ , we define \emph{per user privacy leakage} as
\begin{align*}
  {\rm priv}(\pi, \mathcal{A}_K,k)&=  \max_{i}I(S_{k,i};\widehat{W}| \mathbf{S}_k^{-i}),  % \label{eq:perUsrPrivacyDistribute} 
\\
  {\rm priv}(\pi, \mathcal{A}_K^{(T)},k)&= \max_{i}I(S_{k,i};\widehat{W}^{(T)}| \mathbf{S}_k^{-i}). %\label{eq:perUsrPrivacyFL}
\end{align*}
The above quantities %in \eqref{eq:perUsrPrivacyDistribute} and \eqref{eq:perUsrPrivacyFL} 
should respectively be understood as the privacy leakage of the distributed and federated learning algorithms regarding the data of user $k$. 

%The definition of privacy leakage in \eqref{eqn:def:PrivacyLeakage} becomes unchanged for the centralized case. Now, for the decentralized case in Definition~\ref{def:distributedLearningNoInteraction} and federated learning case in Definition~\ref{def:FederatedLearningAlgorithm} we further define the following quantity: for $k \in \{1, \ldots, K\}$ 
%\begin{align}
%  {\rm priv}(\mu, \mathcal{A}_K,k)&= \max_{i}I(S_{k,i};\widehat{W}| \mathbf{S}_k^{-i}),
%	  \label{eq:perUsrPrivacyDistribute}\\
%  {\rm priv}(\mu, \mathcal{A}_K^{(T)},k)&= \max_{i}I(S_{k,i};\widehat{W}^{(T)}| \mathbf{S}_k^{-i}). \label{eq:perUsrPrivacyFL}
%\end{align}
 
\section{Main Results} 
\label{sec:MainResults}

\subsection{Bounds on Generalization Error and Privacy Leakage in  the Classical (Centralized)  Learning Problem} 

The starting point of our discussion  is the classical learning problem where the server has access to all the data. As the next theorem shows lower and upper bounds on its generalization error, this centralized setup shall be a benchmark in this paper.

\begin{theorem}\label{thm:ClassicalSetting-BoundOnGeneralizationError}   Given a classical (centralized) learning algorithm $\cA$, see Definition~\ref{def:ClassicalLearningAlgorithm}, its generalization error, under the assumption that dataset $\bfS = (S_1, \ldots, S_n) $ is generated by an unknown distribution $P_{\bfS} = \pi^{\otimes n} $, is given by
\begin{align*}
\frac{1}{n}\sum_{i=1}^n \psi_+^{\ast-1} \left(I( S_i;W)\right) &\le {\rm gen}(\pi, \mathcal{A})  \le \frac{1}{n}\sum_{i=1}^n \psi_-^{\ast-1} \left(I( S_i;W)\right)   
%{\rm Sen} ( \pi,  \mathcal{A} ; j )  &\le   \frac{\sigma^2}{2} I( S_j;W| \mathbf{S}^{-j}).  
\end{align*} 
where $\psi_+^{\ast-1} $ and $ \psi_-^{\ast-1} $ are as introduced in Lemma~\ref{lem:AuxiliaryConditionalLemma}.% in Section~\ref{sec:InforTheoreAndPrivacy}.
\end{theorem} 

To give the reader a taste of the generalization error bounds in Theorem~\ref{thm:ClassicalSetting-BoundOnGeneralizationError}, the following lemma shows estimates on generalization error, mutual information and privacy in the classical problem of Gaussian mean estimation.

\begin{lemma}\label{lem:MeanEstiamtionClassical} Let  $K=1$ and  $\pi=\mathcal{N}(\nu, \sigma^2 I_d)$ in Example~\ref{Example:ModelAggreg}.  
%$\psi_+^{\ast-1}(u) = 0$, $\psi_-^{\ast-1}(u)= 2\sqrt{d} \left( 1 + \frac1n  \right) \sigma^2  \sqrt{u} $, ${\rm gen}(\pi, \mathcal{A} ) = \frac{2 \sigma^2 d}{n}$, $ I( S_{k,i}; W)  = \frac{d}{2} \log \left( \frac{n}{n-1}\right)  $, ${\rm priv}( \mathcal{A})=\infty$
\begin{alignat*}{2}
&\psi_+^{\ast-1}(u) = 0 ,   &\psi_-^{\ast-1}(u)= 2\sigma^2\sqrt{d u} \left( 1 + \frac1n  \right)   ,   \\ 
& {\rm gen}(\pi, \mathcal{A} ) = \frac{2 \sigma^2 d}{n}, \qquad   &  I( S_{k,i}; W)  = \frac{d}{2} \log  \frac{n}{n-1}  , \\
&{\rm priv}(\pi, \mathcal{A})=\infty.
\end{alignat*}
\end{lemma}

%{\blue For the Extended Version:  Fig.~\ref{}, using the evaluation in Lemma~\ref{lem:MeanEstiamtionClassical}, compare  the left and the right sides of the bound in Theorem~\ref{thm:ClassicalSetting-BoundOnGeneralizationError}.  } 

 Because the user data is collected at a server, the centralized learning does not provide any sort of privacy guarantees. In this scenario, to protect their own data, a user is therefore needed to add some noise to the data that they provide to the server. This, in turn, reduces the accuracy of the server for the sake of user privacy. As the next section shows, in the distributed learning setting as more and more users join the learning task, the central server will have lesser and lesser ability to infer the data of a single user. Inherently, this shall provide privacy guarantees to each and every participating user. Indeed, perhaps surprisingly, as more users participate in the learning task, not only will the data of a single user be hidden better, but, because more users means better training, the generalization error for the distributed learning algorithm shall also decrease.

\subsection{Bounds on Generalization Error and Privacy Leakage in  the Distributed Learning Problem } 
\label{sec:BOundsDecentralizedCase} 

The following result presents  upper and lower bounds%, in terms of mutual information, 
  { }on the generalization error of a distributed learning algorithm.

\begin{theorem}\label{thm:DistributedCase}    Given a distributed learning algorithm $\cA_K $ with $K$ users, see Definition~\ref{def:distributedLearningNoInteraction}, its generalization error, under the assumption that for each user $k \in \{1,\ldots, K\}$ their dataset $\bfS_k = (S_{k,1}, \ldots, S_{k,n_k} ) $ is generated by an unknown distribution $P_{\bfS_k} = \pi_k^{\otimes n_k} $, is given by  
\begin{align*}
&\frac{1}{n}   \sum_{k=1}^K \sum_{i=1}^{n_k}   \psi_{k+}^{\ast-1}(   I( S_{k,i}; \widehat{W}) ) \le {\rm gen}(\mu; \mathcal{A}_K)   \\ 
&\qquad \le   \frac{1}{n} \sum_{k=1}^K \sum_{i=1}^{n_k} \psi_{k-}^{\ast-1}(   I( S_{k,i}; \widehat{W})). 
\end{align*} 
 where $\mu = \prod_{k=1}^K \pi_k $; and $\psi_+^{\ast-1} $ and $ \psi_-^{\ast-1} $ are as introduced in Lemma~\ref{lem:AuxiliaryConditionalLemma} in Section~\ref{sec:InforTheoreAndPrivacy}.
\end{theorem} 

 The generalization error, mutual information and the provided privacy guarantee in the distributed Gaussian mean estimation problem in Example~\ref{Example:ModelAggreg} are presented in Lemma~\ref{lem:MeanEstiamtionDistributeSetting} below.

\begin{lemma}\label{lem:MeanEstiamtionDistributeSetting}   Let $\pi_k=\mathcal{N}(\nu, \sigma_k^2 I_d)$ in Example~\ref{Example:ModelAggreg}. Then, 
\begin{align*}
& \psi_{k+}^{\ast-1}(u) = 0 , \qquad  \psi_{k-}^{\ast-1}(u) = 2 \sqrt{d u}\left( \sum_{i=1}^K \frac{n_i \sigma_i^2 }{n^2} + \sigma_k^2 \right)   ,\\
& {\rm gen}(\mu, \mathcal{A}_K  )  =  \sum_{i= 1}^K \frac{2d n_i\sigma_i^2}{n^2} , \
\\ &  
I( S_{k,i}; \widehat{W})  = \frac{d}{2} \log \bigg(1 + \frac{\sigma_k^2 }{\sum_{i=1}^K n_i \sigma_i^2 - \sigma_k^2} \bigg) , \\
& {\rm priv} (\pi, \mathcal{A}_K , k)  = \frac{d}{2} \log \bigg(1+    \frac{ \sigma_k^2}{ \sum_{i=1: i \neq k}^K n_i \sigma_i^2 }\bigg).
\end{align*}
\end{lemma}

 Although too simple for any practical purpose,  it is still illuminating to consider the   case  where each user have equal amount of data which are i.i.d., in this case:
\begin{align*}
	{\rm gen}(\mu, \mathcal{A}_K  )  =   \frac{2d  \sigma^2}{ K n}, \qquad 
	{\rm priv} = \frac{d}{2} \log \left(1+    \frac{ 1}{ n(K-1) }\right).
\end{align*}
 There are several key observations to make in this basic example.  The simplest is that  the generalization error  in this distributed  learning problem corresponds to  that of the classical learning problem with a dataset of size $Kn$. This means that, as far as the generalization error is concerned, distributing the learning task causes no loss in optimality.  More importantly,   as opposed to the classical learning  case, the privacy leakage here in this example is finite. In fact,  for each participating user the privacy leakage increases, fairly fast, as the number of users in the system increases, which means that multi-user nature of distributed learning provides an inherent privacy to each of its users.

\subsection{Bounds on Generalization Error and Privacy Leakage in the Federated Learning Problem } 

Using the same technique applied in classical and distributed learning settings, the following result gives upper and lower bounds, in term of mutual information, on the generalization error at time $t$ of a federated learning algorithm.

\begin{theorem}\label{thm:FederateLearning}   
Given a federated learning algorithm $\cA^{(T)}_K$ with $K$ users communicating with the server a total of $T$ rounds, see Definition~\ref{def:FederatedLearningAlgorithm}, its generalization error at time $t$, under the assumption that for each user $k\in \{1, \ldots, K\} $ their most up-to-date dataset $\bfS_k^{(t)} = (S_{k,1}, \ldots, S_{k,n_k}) $ is generated by an unknown distribution $P_{\bfS_k^{(t)}}  = \pi_k^{\otimes n_k } $, is given by
\begin{align*}
&- \bbE\left[ \frac{1}{n^{(t)}} \sum_{k\in \cI^{(t)}} \sum_{i=1}^{n_k}\psi_+^{*-1}\left(I(S^{(t)}_{k,i} ; \widehat W^{(t)})\right)\right]\notag 
%\\ &\qquad 
\le {\rm gen}^{(t)}(\mu;    \mathcal{A}_K^{(T)})   \\
 &\qquad \le \bbE\left[  \frac{1}{n^{(t)}} \sum_{k\in \cI^{(t)}} \sum_{i=1}^{n_k}\psi_-^{*-1}\left(I(S^{(t)}_{k,i} ; \widehat W^{(t)})\right)\right]
\end{align*}
where the expectations are with respect to the $\cI^{(t)}$.
\end{theorem} 

Following lemma illustrates the generalization error bound and privacy leakage in federated learning algorithm.

\begin{lemma} In Example~\ref{Example:ModelAggregFL} with the simplifying assumptions that for each user $n_k = n$, $\sigma_k^2 = \sigma^2 $, there are $K_a^{(t)} = K_a $ active users at all times $t$ which are uniformly chosen, and $\pi = \cN (\nu, \sigma^2 I_d )  $ for each user,
\begin{align*}
&\psi_{k+}^{\ast-1}(u) = 0, \qquad \quad \ \psi_{k-}^{\ast-1}(u) = 2 \sqrt{d} \left(1+ \frac{1}{tnK_a}\right)\sigma^2  \sqrt{u},\\
&{\rm gen}(\mu;    \mathcal{A}_K^{(t)}) = \frac{2d\sigma^2}{tnK}, 
\\  &  
I(S_{k^*j^*}(\mfrI_t ); \widehat W) = \frac{d}{2} \log \left(1+  \frac{1}{tnK_a -1 }\right ) .
%  \\ {\rm priv}(  \mathcal{A}_K , k)   & =???
\end{align*}
\end{lemma}

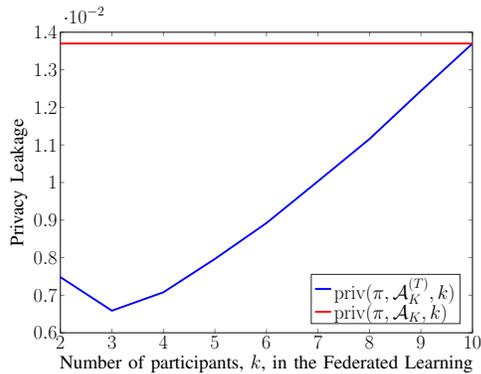
\begin{figure}[h!]
\centering
\resizebox{0.35\textwidth}{!}{%
\begin{tikzpicture}

\begin{axis}[%
width=6.028in,
height=4.4in,
at={(1.011in,0.642in)},
scale only axis,
axis line style = thick,
xmin=2,
xmax=10,
ymin=0.006,
ymax=0.014,
xtick={2, 3, ..., 9, 10},
xlabel = {\huge Number of participants, $k$, in the Federated Learning}  ,
ylabel = {\huge Privacy Leakage},
axis background/.style={fill=white},
tick label style={font=\huge},
legend style={at={(0.97,0.03)}, anchor=south east, legend cell align=left, align=left, draw=white!15!black}
]
\addplot [color=blue, line width=2pt]
  table[row sep=crcr]{%
2	0.00748137999581369\\
3	0.00658827178318392\\
4	0.00707875438431711\\
5	0.00796726177860729\\
6	0.00892106039769886\\
7	0.0100339242174204\\
8	0.011157110522785\\
9	0.0124432379704582\\
10	0.0136994870940572\\
};
\addlegendentry{\huge ${\rm priv}(\pi, \cA^{(T)}_K, k ) $}

\addplot [color=red, line width=2pt]
  table[row sep=crcr]{%
2	0.0136994870940572\\
3	0.0136994870940572\\
4	0.0136994870940572\\
5	0.0136994870940572\\
6	0.0136994870940572\\
7	0.0136994870940572\\
8	0.0136994870940572\\
9	0.0136994870940572\\
10	0.0136994870940572\\
};
\addlegendentry{\huge ${\rm priv}(\pi, \cA_K, k ) $}

\end{axis}
\end{tikzpicture}%
}
\caption{Comparison of privacy leakages in distributed and federated learning. } \label{fig:comparisonbtwDistrvsFederated}
\end{figure}

While we cannot analytically find the privacy leakage in this example, a comparison plot between the distributed learning and federated learning cases can be seen in Figure~\ref{fig:comparisonbtwDistrvsFederated}. Note that there are $K=10$  users in the distributed learning case. The $x$-axis shows the number of active users $K_a$ in the federated learning setting, observe that if all 10 of the users are active in the federated case, we get the same amount of privacy leakage. However, if at this round, there are less than 10 active users, federated learning algothim allows less privacy leakage as compared to its distributed counterpart.

%%%%%%%%%%%%%%%%%%%%%%%%%%%%%%
\bibliography{refs}

\begin{thebibliography}{10}
\providecommand{\url}[1]{#1}
\csname url@samestyle\endcsname
\providecommand{\newblock}{\relax}
\providecommand{\bibinfo}[2]{#2}
\providecommand{\BIBentrySTDinterwordspacing}{\spaceskip=0pt\relax}
\providecommand{\BIBentryALTinterwordstretchfactor}{4}
\providecommand{\BIBentryALTinterwordspacing}{\spaceskip=\fontdimen2\font plus
\BIBentryALTinterwordstretchfactor\fontdimen3\font minus
  \fontdimen4\font\relax}
\providecommand{\BIBforeignlanguage}[2]{{%
\expandafter\ifx\csname l@#1\endcsname\relax
\typeout{** WARNING: IEEEtran.bst: No hyphenation pattern has been}%
\typeout{** loaded for the language `#1'. Using the pattern for}%
\typeout{** the default language instead.}%
\else
\language=\csname l@#1\endcsname
\fi
#2}}
\providecommand{\BIBdecl}{\relax}
\BIBdecl

\bibitem{mcmahan2016communication}
H.~B. McMahan, E.~Moore, D.~Ramage, S.~Hampson \emph{et~al.},
  ``Communication-efficient learning of deep networks from decentralized
  data,'' \emph{preprint arXiv:1602.05629}, 2016.

\bibitem{shamir2014distributed}
O.~Shamir and N.~Srebro, ``Distributed stochastic optimization and learning,''
  in \emph{Proc. 52nd Allerton Conf. Commun., Control and Comp.}, 2014, pp.
  850--857.

\bibitem{russo2015much}
D.~Russo and J.~Zou, ``How much does your data exploration overfit?
  {C}ontrolling bias via information usage,'' \emph{preprint arXiv:1511.05219},
  2015.

\bibitem{jiao2017dependence}
J.~Jiao, Y.~Han, and T.~Weissman, ``Dependence measures bounding the
  exploration bias for general measurements,'' in \emph{Proc. of IEEE Int.
  Symp. Inform. Theory (ISIT)}, 2017, pp. 1475--1479.

\bibitem{xu2017information}
A.~Xu and M.~Raginsky, ``Information-theoretic analysis of generalization
  capability of learning algorithms,'' in \emph{Proc. of Neural Information
  Processing Systems}, 2017, pp. 2524--2533.

\bibitem{asadi2018chaining}
A.~Asadi, E.~Abbe, and S.~Verd{\'u}, ``Chaining mutual information and
  tightening generalization bounds,'' in \emph{Proc. of Neural Information
  Processing Systems}, 2018, pp. 7234--7243.

\bibitem{bu2019tightening}
Y.~Bu, S.~Zou, and V.~V. Veeravalli, ``Tightening mutual information based
  bounds on generalization error,'' \emph{preprint arXiv:1901.04609}, 2019.

\bibitem{pensia2018generalization}
A.~Pensia, V.~Jog, and P.-L. Loh, ``Generalization error bounds for noisy,
  iterative algorithms,'' in \emph{Proc. of IEEE Int. Symp. Inform. Theory
  (ISIT)}, 2018, pp. 546--550.

\bibitem{negrea2019information}
J.~Negrea, M.~Haghifam, G.~K. Dziugaite, A.~Khisti, and D.~M. Roy,
  ``Information-theoretic generalization bounds for {SGLD} via data-dependent
  estimates,'' in \emph{Proc. of Neural Information Processing Systems}, 2019,
  pp. 11\,013--11\,023.

\bibitem{raginsky2016information}
M.~Raginsky, A.~Rakhlin, M.~Tsao, Y.~Wu, and A.~Xu, ``Information-theoretic
  analysis of stability and bias of learning algorithms,'' in \emph{Proc. of
  IEEE Inf. Theory Workshop}, 2016, pp. 26--30.

\bibitem{wang2019information}
H.~Wang, M.~Diaz, J.~C.~S. Santos~Filho, and F.~P. Calmon, ``An
  information-theoretic view of generalization via {W}asserstein distance,'' in
  \emph{Proc. of IEEE Int. Symp. Inform. Theory (ISIT)}, 2019, pp. 577--581.

\bibitem{esposito2019generalization}
A.~R. Esposito, M.~Gastpar, and I.~Issa, ``Generalization error bounds via
  {R\'enyi}, $f$-divergences and maximal leakage,'' \emph{preprint
  arXiv:1912.01439}, 2019.

\bibitem{dwork2006calibrating}
C.~Dwork, F.~McSherry, K.~Nissim, and A.~Smith, ``Calibrating noise to
  sensitivity in private data analysis,'' in \emph{Proc. Theory of Crypt.
  Conf.}, 2006, pp. 265--284.

\bibitem{cuff2016differential}
P.~Cuff and L.~Yu, ``Differential privacy as a mutual information constraint,''
  in \emph{Proc. of the ACM SIGSAC Conf. on Comp. and Commun. Security}, 2016,
  pp. 43--54.

\bibitem{yang2015bayesian}
B.~Yang, I.~Sato, and H.~Nakagawa, ``Bayesian differential privacy on
  correlated data,'' in \emph{Proc. of the ACM SIGMOD International Conf. on
  Management of Data}, 2015, pp. 747--762.

\bibitem{li2019federated}
T.~Li, A.~K. Sahu, A.~Talwalkar, and V.~Smith, ``Federated learning:
  {C}hallenges, methods, and future directions,'' \emph{preprint
  arXiv:1908.07873}, 2019.

\bibitem{kairouz2019advances}
P.~Kairouz, H.~B. McMahan, B.~Avent, A.~Bellet, M.~Bennis, A.~N. Bhagoji,
  K.~Bonawitz, Z.~Charles, G.~Cormode, R.~Cummings \emph{et~al.}, ``Advances
  and open problems in federated learning,'' \emph{preprint arXiv:1912.04977},
  2019.

\end{thebibliography}
\bibliographystyle{IEEEtran}

\end{document}